\documentclass{article}
\usepackage{graphicx}
\usepackage[font=small]{caption}
\usepackage{wrapfig}

\usepackage[final]{nips_2016}

\usepackage[utf8]{inputenc} 
\usepackage[T1]{fontenc}    
\usepackage{hyperref}       
\usepackage{url}            
\usepackage{booktabs}       
\usepackage{amsfonts}       
\usepackage{nicefrac}       
\usepackage{microtype}      

\title{The Use of Autoencoders for Discovering Patient Phenotypes}

\author{
  Harini Suresh \\
  MIT CSAIL\\
  Cambridge, MA 02139 \\
  \texttt{hsuresh@mit.edu@mit.edu} \\
  \And
  Peter Szolovits \\
  MIT CSAIL\\
  Cambridge, MA 02139 \\
  \texttt{psz@mit.edu}
  \And
  Marzyeh Ghassemi \\
  MIT CSAIL\\
  Cambridge, MA 02139 \\
  \texttt{mghassem@mit.edu}
}

\begin{document}

\maketitle

\begin{abstract}
We use autoencoders to create low-dimensional embeddings of underlying patient phenotypes that we hypothesize are a governing factor in determining how different patients will react to different interventions. We compare the performance of autoencoders that take fixed length sequences of concatenated timesteps as input with a recurrent sequence-to-sequence autoencoder.  We evaluate our methods on around 35,500 patients from the latest MIMIC III dataset from Beth Israel Deaconess Hospital.  

\end{abstract}
\section{Introduction }

Intensive Care Units (ICUs) are high-cost, limited-resource environments where quick and accurate decision-making is extremely valuable.  However, most decision-making is often made in settings of high uncertainty and based just on the clinician's prior experience.  With the rapid rise in Electronic Health Records (EHRs) available for analysis, data-driven models are well-suited to aid physicians in making decisions about when patients should be treated with or weaned off certain interventions [1].  

Achieving high-quality care depends on a robust understanding of the patient's underlying acuity throughout time [2].  Traditional measures of acuity are often based on mortality evaluated at a single endpoint [3, 4], or on static scores such as SAPS that don't take into account evolving clinical information [5, 6].  

We aim to create richer representations of patient health with the end goal of predicting actionable interventions.  The efficacy of interventions can drastically vary from patient to patient,  and unnecessarily administering an intervention can be harmful and expensive [7].  A model of patient health that is able to capture complex relationships in physiological signals is key to accurately predicting onset/weaning of interventions for different patients and necessary for successful personalized medicine. 

This type of patient phenotyping is challenging because robust representations of human physiology are complicated, and contain many non-obvious dependencies between observed measurements.  Moreover, modelling evolving clinical information requires using timeseries data, but this data is often varying-length, irregularly sampled or has missing values.  Previously, multitask gaussian processes have been tested for modelling patient acuity but only in Traumatic Brain Injury (TBI) patients [8] or only using longitudinal billing data [2].  

Our approach uses autoencoders for physiological timeseries signal reconstruction.  Autoencoders deep neural networks where the target values are the same as the input values, and the hidden layer(s) compress the inputs into a lower dimensional embedding. Since this embedding tries to reconstruct the original input, it must capture fundamental features about the input timeseries, and can be used as a measure of underlying patient acuity.  Feature learning in this approach is entirely unsupervised, so unlike traditional acuity scores it is not limited by a manually-defined feature space.  Furthermore, recurrent autoencoders are able to model signals of varying-length and are robust to missing data due to the ability of Long-Short Term Memory (LSTM) cells to forget unimportant inputs.  

Using autoencoders allows us to extract the low-dimensional embedding and use it in further work for predicting interventions.

\section{Background}

\subsection{Autoencoders}

Previous work has attempted to use autoencoders on random 30-day patches of input vectors to learn underlying patient phenotypes [9], and has shown that they can extract relevant and useful features about patients.    

It is also valuable to consider signals that occur throughout a patient's stay, since there are often long delays between when relevant signals are expressed [10]. 

Sequence autoencoders take in measured signals one timestep at a time into a layer of LSTM (Long-Short Term Memory) cells and produce a fixed-length embedding.  This embedding is then used as input to another layer of LSTM cells that try to predict the original input sequence. 

LSTM cells are used because of their ability to effectively model varying-length timeseries data and capture long-term dependencies.  LSTMs have achieved state-of-the-art results in many different natural language processing applications from machine translation [11] to dialogue systems [12] to image captioning [13].   They are well-suited to modeling clinical data because evidence of certain conditions may be spread apart over several hours or days, and important symptoms may present early on in a patient's stay.  

Sequence autoencoders using LSTM cells were inspired by the success of general sequence-to-sequence models applied to machine translation.  They were recently used as an initialization step for recurrent neural networks for text classification [14], but have not been applied to the clinical space.

\section{Methods}

\subsection{Features}

Features were extracted from the Multiparameter Intelligent Monitoring in Intensive Care (MIMIC III) Database [15]. Since 2001, the MIMIC database has been built up and maintained by the Laboratory of Computational Physiology at the Massachusetts Institute of Technology, Beth Israel Deaconess Medical Center, and Philips Healthcare, with support from the National Institute of Biomedical Imaging and Bioinformatics. The database contains general patient information (ICD-9 codes, demographics, room tracking), physiological signals (vital metrics, SAPS), medications (IV meds, provider order entry data), lab tests (chemistry, imaging), fluid balance (in- take, output), and notes (discharge summary, nursing progress reports).  The most recent version of MIMIC III contains data from around 38,600 adults, comprising over 58,000 hospital admissions, from 2001-2012.  The MIMIC dataset is notable because it is publicly available for free use, encompasses a large and diverse set of patients, and contains numerous high resolution features for each patient.  

We use data from from the first ICU stay of patients in the Medical Care Unit (MICU), Cardiac Care Unit (CCU), Cardiac Surgery Recovery Unit (CSRU), Surgical Intensive Care Unit (SICU), and Trauma Surgical Intensive Care Unit (TSICU).  We only look at patients older than 15 and who were in the ICU for between 12 and 2000 hours.   This totals 35,554 unique patients/ICU stays. We extract 30 physiological features for each patient as timeseries spanning their entire stay.  Measurements are given timestamps that are rounded to the nearest hour, and if an hour has multiple measurements for a signal those are averaged.

Since there are many missing values, we first backfill using any present values the patient has, and then fill in remaining missing values with the mean value for that variable across all patients.  

The data is split into training/validation/testing sets with a 70/15/15 split, stratified on in-hospital mortality in order to have a spectrum of patient severity in both the train and test sets.  

\subsection{Autoencoders}

We test the ability of a simple autoencoder with a single hidden layer, an autoencoder with two hidden layers, and a sequence autoencoder to reconstruct the input (Figure 1).  We also compare the performance of these models over inputs of different interval lengths, specifically 4, 16, 32 and 64 hours.  We train on mini-batches of 128 samples with early stopping to determine the number of epochs.    
\begin{figure}
\includegraphics[width=420px]{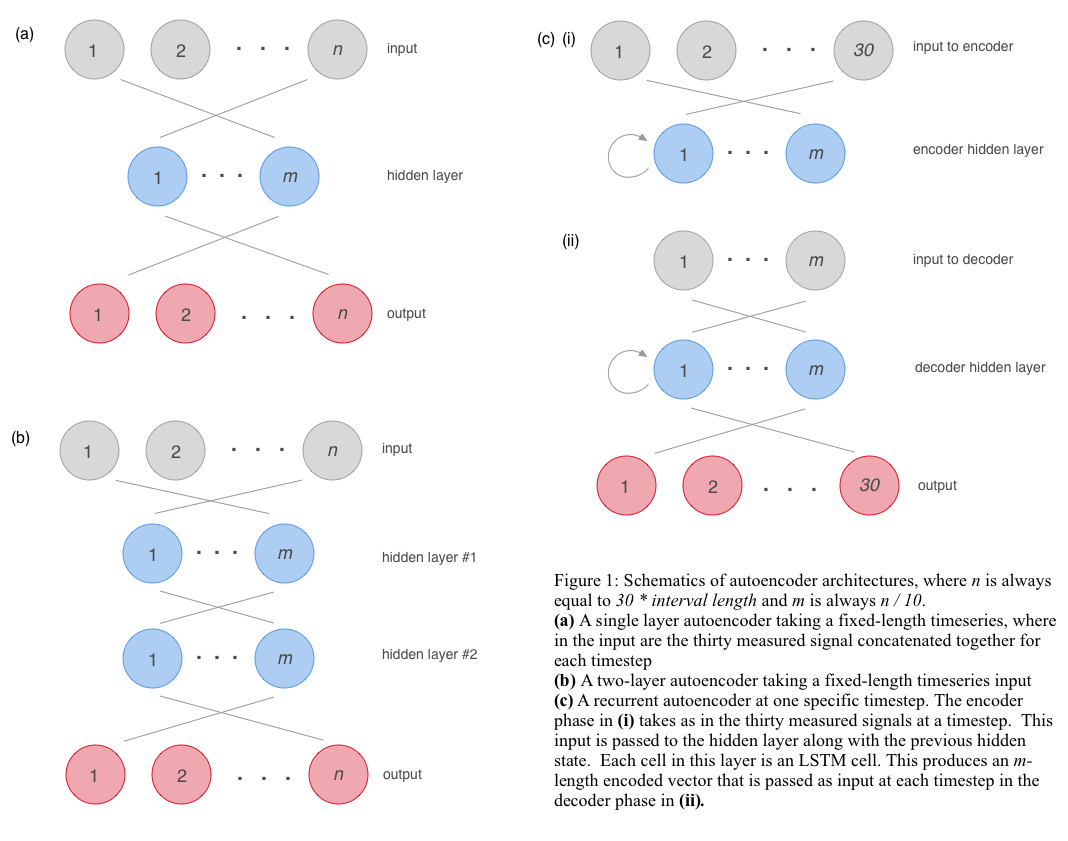}
\end{figure}

For the fixed-length input autoencoders, we concatenate all 30 features for each hour throughout the given interval length.  We use an embedding size equal to the total number of input values divided by 10 to achieve a compression factor of 10x. All hidden layers use a ReLU activation function, and the output layer uses a sigmoidal activation function.  

\section{Results}

\setcounter{figure}{1}
\begin{figure}
\centering
\begin{minipage}{.5\textwidth}
  \centering
  \captionsetup{width=.9\linewidth}
  \includegraphics[width=.99\textwidth]{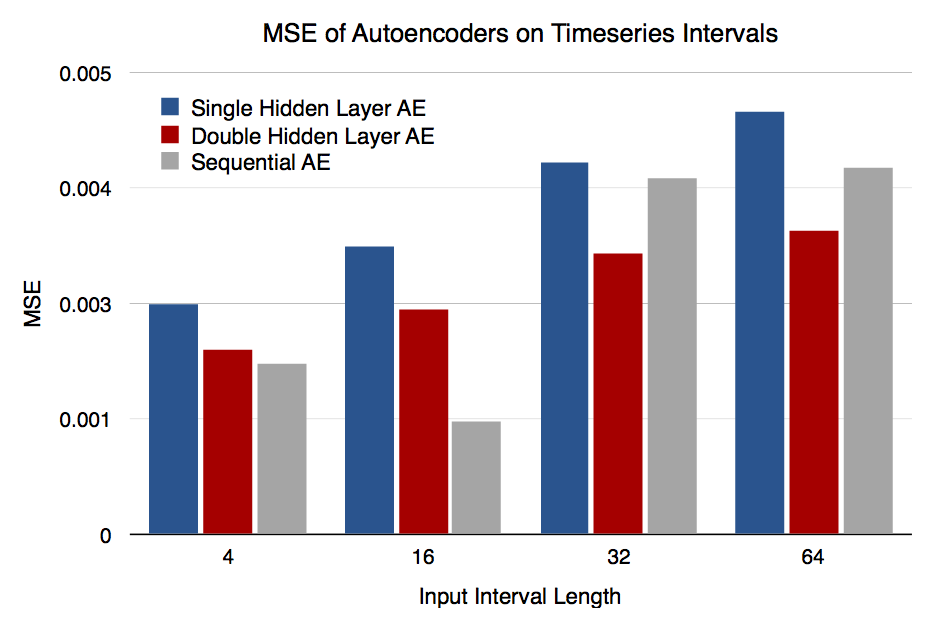}
  \caption[width=0.8\textwidth]{Performance of autoencoders on reconstructing timeseries input of various lengths.}
\end{minipage}%
\begin{minipage}{.5\textwidth}
  \centering
  \captionsetup{width=.9\linewidth}
  \includegraphics[width=.99\textwidth]{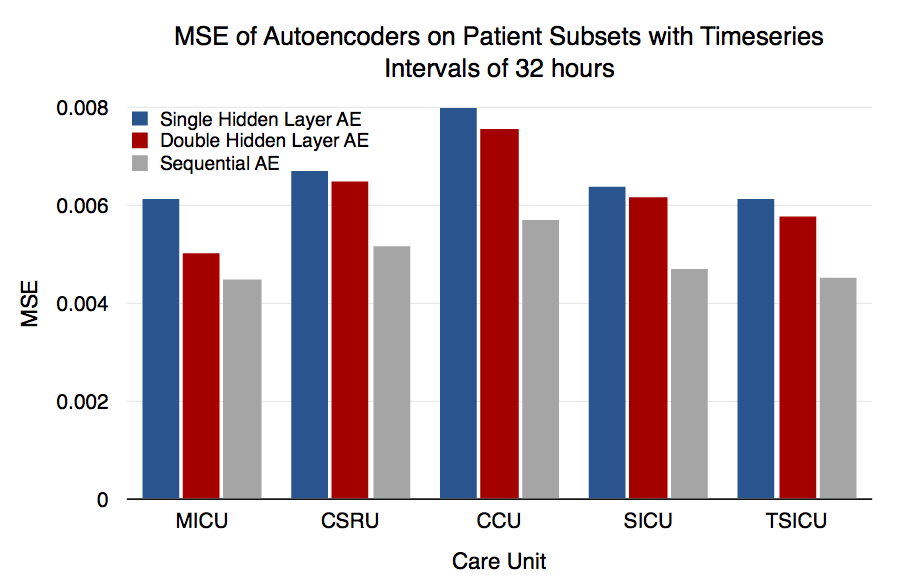}
  \caption[width=0.8\textwidth]{Performance of autoencoders on patient population subsets with intervals of 32 hours.}
\end{minipage}
\end{figure}
We evaluate the performance of each autoencoder by taking the mean squared error (MSE) between the predicted sequence of values and the true sequence of values.  The sequential autoencoder with one LSTM layer achieves a lower MSE than the single-layer fixed length autoencoder on all interval lengths, but varies in comparison to the double-layer fixed length autoencoder (Figure 2).

We also show that reconstructing input timeseries with autoencoders is fairly robust to stratifications in population subsets.  We run the autoencoders on intervals of 32 hours with patient subsets stratified by care unit.  MSEs are higher than when the autoencoders were trained on the entire patient population, but less than 0.08 in all cases, even though the training sets are much smaller (Figure 3).  On these smaller subsets of patients, the sequence autoencoder appears to be able to generalize better with smaller amounts of training data and does better in all cases. 

\begin{wrapfigure}{l}{0.6\textwidth}
 \captionsetup{width=.5\textwidth}
\centering
\includegraphics[width=0.58\textwidth]{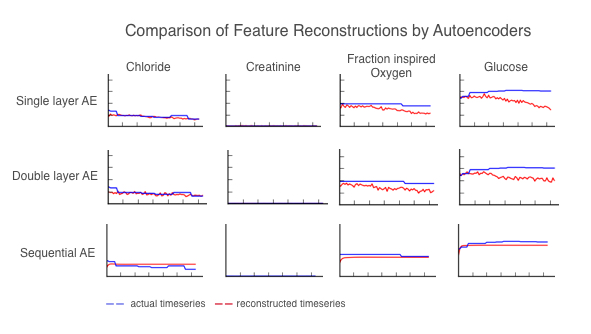}
\caption{Examples of feature reconstructions for a single patient for an interval of 64 hours.}
\end{wrapfigure}

The reconstructed features from the sequence autoencoder are less noisy and more robust to varying-length sequences.  The reconstructed feature values from the fixed-length autoencoder often start trailing towards zero near the end of the sequence (Figure 4).  This could be due to the fact that sequences are zero-padded at the end for patients with fewer hours of data than the interval length, and the point at which a patient's real data ends is harder to determine when the sequence is fed in as one long concatenated vector.   

\section{Future Work}

We plan to compare the performance of patient embeddings from different autoencoder structures to predict onset of and weaning off of dialysis.  We will compare its performance on its own and as an additional feature matrix concatenated to the raw feature values.  

It will also be interesting to experiment with deeper sequential autoencoder structures, or to use bidirectional LSTM cells in the hidden layers to better reconstruct inputs.

\section*{References}

\small

[1] Vincent JL. Critical care-where have we been and where are we going? \emph{Critical Care}. 2013;17(Suppl 1):S2.

[2] Che, Zhengping, et al. "Deep computational phenotyping." \emph{Proceedings of the 21th ACM SIGKDD International Conference on Knowledge Discovery and Data Mining.} ACM, 2015.

[3] Caballero Barajas KL, Akella R. Dynamically modeling patient’s health state from electronic medical records: a time series approach. In: \emph{Proceedings of the 21th ACM SIGKDD International Conference on Knowledge Discovery and Data Mining}. New York: ACM; 2015: 69–78.

[4] Ghassemi M, Naumann T, Doshi-Velez F, et al. Unfolding physiological state: Mortality modelling in intensive care units. In:\emph{ Proceedings of the 20th ACM SIGKDD International Conference on Knowledge Discovery and Data Mining}. ACM, 2014:75–84.

[5] Nassar, Antonio Paulo, et al. “Caution when using prognostic models: a prospective comparison of 3 recent prognostic models.” \emph{Journal of critical care} 27.4 (2012): 423-e1.

[6] Arabi, Yaseen, et al. “Assessment of six mortality prediction models in patients admitted with severe sepsis and septic shock to the intensive care unit: a prospective cohort study.” \emph{Critical care} 7.5 (2003): R116.

[7] D’Aragon F, Belley-Cote EP, Meade MO, et al. Blood pressure targets for vasopressor therapy: a systematic review. \emph{Shock} 2015;43:530–9.

[8] Ghassemi, Marzyeh, et al. "A multivariate timeseries modeling approach to severity of illness assessment and forecasting in ICU with sparse, heterogeneous clinical data." \emph{Proceedings of the...AAAI Conference on Artificial Intelligence. }AAAI Conference on Artificial Intelligence. Vol. 2015. NIH Public Access, 2015.

[9] Lasko, Thomas A., Joshua C. Denny, and Mia A. Levy. "Computational phenotype discovery using unsupervised feature learning over noisy, sparse, and irregular clinical data." \emph{PloS} one 8.6 (2013): e66341.

[10] Lipton, Zachary C., et al. "Learning to Diagnose with LSTM Recurrent Neural Networks." arXiv preprint arXiv:1511.03677 (2015).

[11] Bahdanau, Dzmitry, Kyunghyun Cho, and Yoshua Bengio. "Neural machine translation by jointly learning to align and translate." arXiv preprint arXiv:1409.0473 (2014).

[12] Wen, Tsung-Hsien, et al. "Semantically conditioned lstm-based natural language generation for spoken dialogue systems." arXiv preprint arXiv:1508.01745 (2015).
APA	

[13] Vinyals, Oriol, et al. "Show and tell: A neural image caption generator." \emph{Proceedings of the IEEE Conference on Computer Vision and Pattern Recognition.} 2015.

[14] Dai, Andrew M., and Quoc V. Le. "Semi-supervised sequence learning." \emph{Advances in Neural Information Processing Systems. }2015.

[15] Saeed, Mohammed, Mauricio Villarroel, Andrew T. Reisner, Gari Clif- ford, Li-Wei Lehman, George Moody, Thomas Heldt, Tin H. Kyaw, Benjamin Moody, and Roger G. Mark. “Multiparameter Intelligent Moni- toring in Intensive Care II (MIMIC-II): A public-access intensive care unit database.” 2011; 39:952-960. DOI: 10.1097/CCM.0b013e31820a92c6.

\end{document}